\documentclass[lettersize,journal]{IEEEtran}
\usepackage{amsmath,amsfonts}
\usepackage{algorithmic}
\usepackage{algorithm}
\usepackage{array}
\usepackage{textcomp}
\usepackage{stfloats}
\usepackage{url}
\usepackage{verbatim}
\usepackage{graphicx}
\usepackage{cite}
\usepackage{color}
\usepackage{caption}
\usepackage{subcaption}
\usepackage{enumitem}
\usepackage{multicol}
\usepackage{multirow}
\usepackage{pifont}         

\newlength\savewidth\newcommand\shline{\noalign{\global\savewidth\arrayrulewidth
		\global\arrayrulewidth 1pt}\hline\noalign{\global\arrayrulewidth\savewidth}}

\begin{document}

\title{MSight: An Edge-Cloud Infrastructure-based Perception System for Connected Automated Vehicles}

\author{Rusheng Zhang$^*$, Depu Meng$^*$,~\IEEEmembership{Member,~IEEE},
Shengyin Shen, Zhengxia Zou, \\
Houqiang Li,~\IEEEmembership{Fellow,~IEEE}
, Henry X. Liu,~\IEEEmembership{Member,~IEEE}


\thanks{
This work was supported by Mcity of the University of Michigan.
\textit{Corresponding author: 
Henry X. Liu.}}

\thanks{R. Zhang is with the Department
of Civil and Environmental Engineering, University of
Michigan, Ann Arbor, MI, 48109, USA (email: rushengz@umich.edu)}

\thanks{D. Meng is with the Department
of Civil and Environmental Engineering, University of
Michigan, Ann Arbor, MI, 48109, USA (email: depum@umich.edu)
}

\thanks{S. Shen is with the University of Michigan 
Transportation Research
Institude, 2901 Baxer Rd, Ann Arbor, MI, 48109, USA.
(email: shengyin@umich.edu)}

\thanks{Z. Zou is with the Department of Guidance, Navigation and Control, School of Astronautics, Beihang University, Beijing 100191, China. (email: zhengxiazou@buaa.edu.cn)}

\thanks{
Houqiang Li is with
the CAS Key Laboratory of
Technology in Geo-Spatial Information Processing and Application System,
Department of Electronic Engineering and Information Science, 
University of Science and Technology of China,
Hefei, China (email:
lihq@ustc.edu.cn)}

\thanks{H. X. Liu is with the Department
of Civil and Environmental Engineering, University of
Michigan, Ann Arbor, MI, 48109, USA.
H. X. Liu is also with Mcity, University of Michigan,
Ann Arbor, MI, 48109, USA (email: henryliu@umich.edu)}

\thanks{$^*$Equal contribution.}

}
\maketitle

\begin{abstract}
As vehicular communication and networking technologies continue to advance, infrastructure-based roadside perception emerges as a pivotal tool for connected automated vehicle (CAV) applications. Due to their elevated positioning, roadside sensors, including cameras and lidars, often enjoy unobstructed views with diminished object occlusion. This provides them a distinct advantage over onboard perception, enabling more robust and accurate detection of road objects. This paper presents MSight, a cutting-edge roadside perception system specifically designed for CAVs. MSight offers real-time vehicle detection, localization, tracking, and short-term trajectory prediction. Evaluations underscore the system's capability to uphold lane-level accuracy with minimal latency, revealing a range of potential applications to enhance CAV safety and efficiency. Presently, MSight operates 24/7 at a two-lane roundabout in the City of Ann Arbor, Michigan.
\end{abstract}


\section{Introduction}
\begin{figure}
    \centering
    \includegraphics[width=\columnwidth]{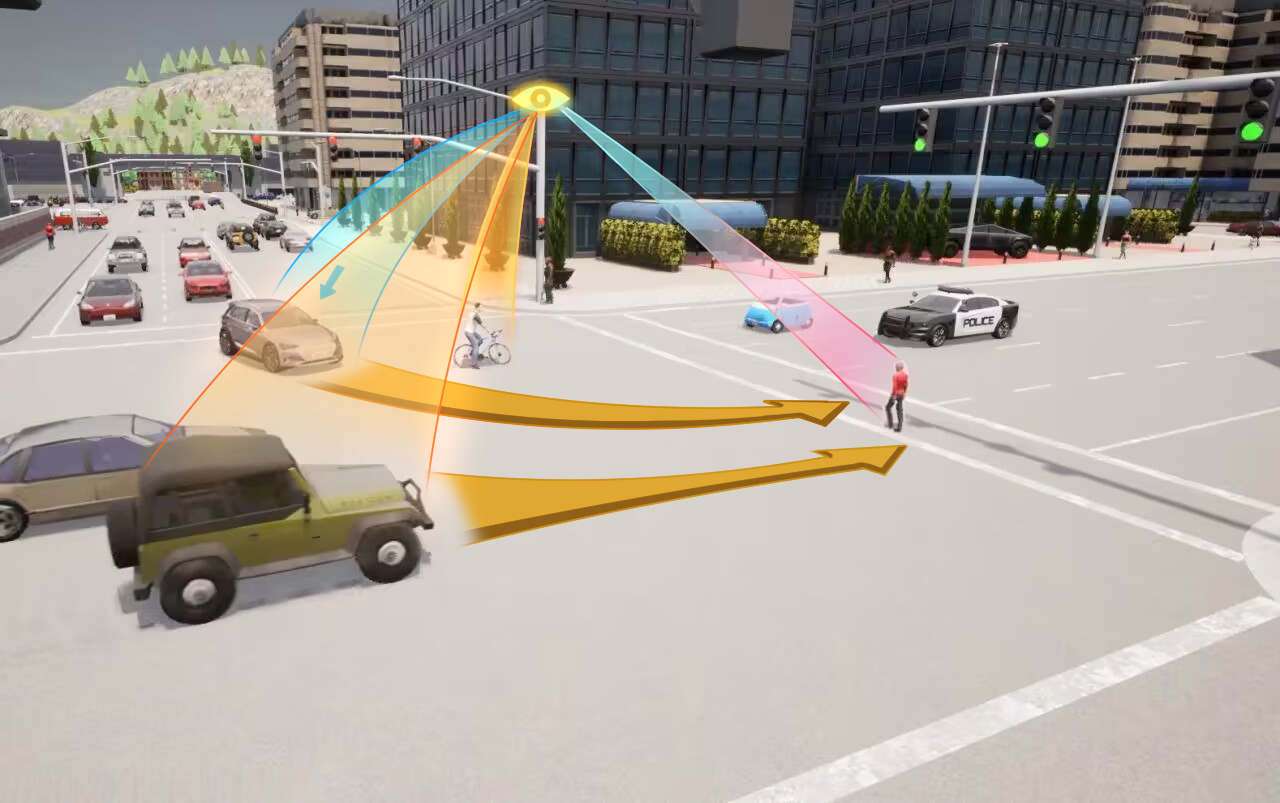}
    \caption{An illustrative figure of cooperative perception. The roadside sensor detects vehicles on the road, and the perception results are forwarded to connected vehicles via roadside radio.}
    \label{fig:illustrative}
\end{figure}

With the rapid development in vehicular communication and networking technologies, infrastructure-based roadside vehicle detection has become viable for connected automated vehicle (CAV) applications. Roadside perception systems can be more accurate, reliable, and with better coverage than vehicle onboard perception systems due to smaller background variations and elevated positions. Connected vehicles can receive the roadside perception results via vehicle-to-infrastructure (V2I) and vehicle-to-everything (V2X) communication with short latency to improve the perception quality cooperatively, as shown in Figure \ref{fig:illustrative}.

While many current automated vehicle solutions rely on onboard perception systems, their limitations become increasingly evident as the demands for more complex perception tasks grow. Occlusion stands out as a significant drawback of onboard perception, potentially leading to safety-critical situations~\cite{schwall2020waymo}. Computational power onboard also constrains the application of resource-intensive perception algorithms. In contrast, roadside perception systems can employ high-performance computing hardware, facilitating more sophisticated algorithms and minimizing inference latency. Although they bear similarities to surveillance and monitoring systems, roadside perception systems for connected and automated vehicles distinctly demand high localization and tracking accuracy coupled with low latency.

Roadside perception systems also have significant impacts in the realm of vehicular communication. Through V2V communications, vehicles broadcast Basic Safety Messages (BSMs) as dictated by the SAE 2735 protocol \cite{kenney2011dedicated}. These messages are broadcast either via the 802.11p channel or through C-V2X technology \cite{chen2017vehicle, chen2020vision}. These broadcasts relay crucial safety data, such as location, heading, brake status, and more, which neighboring vehicles can utilize. Yet, the uptake of this technology is sluggish, largely due to a classic 'chicken-and-egg' dilemma. Its sparse adoption offers minimal advantages to early adopters, leading to hesitancy among manufacturers, which perpetuates the cycle of slow adoption. To overcome this issue, proxy solutions via V2I communications are proposed to boost penetration rate at certain locations. Roadside radios, termed as Roadside Units (RSUs), encode these detection results, disseminating them to all CAVs in the vicinity. This creates an effect synonymous with high adoption within chosen regions. An initial method involves encoding perception data to resemble broadcasts from vehicles, referred to as Proxy-BSM in the USA or Proxy-CAM (cooperative awareness message) in Europe and Japan \cite{kitazato2016proxy}. More recently, the SAE J3224 standard has been put forward, offering a structured approach using Sensor Data Sharing Messages (SDSMs) \cite{sae2019v2x}.

Currently, roadside perception systems for 
cooperative driving is still in its infancy and many reported results are still in lab or early testing stage. The existing works are preliminary and far from real-world deployment. In this paper, we introduce MSight, a full-stack end-to-end roadside perception system. MSight stands out from existing solutions by aiming to achieve high accuracy and low latency that meets the requirements of CAV applications.

There are several aspects of MSight that make it an imperative step towards a production-ready cooperative perception system. First of all, MSight is a full-stack end-to-end cooperative perception system that contains not only a state-of-the-art perception algorithm but also a complete communication pipeline with CAVs and the cloud data center. Furthermore, the system has been deployed and operated 24/7 on the roadside in a production environment. Finally, this system has been extensively tested with CAVs, and the field test results are presented in this paper.
It is anticipated that MSight will accelerate the deployment of CAVs and speed up the adoption of V2X communications technologies. Despite its primary focus on CAV applications, the edge-cloud architecture of MSight makes it possible to extend its functionalities in a wide range of different applications.

\section{Related Works}
\label{s:relatedWorks}

Sensor-based roadside surveillance systems date back to 1986. Initially, these systems were used to detect abnormal vehicle behavior \cite{datondji2016survey}. The development of roadside surveillance/perception systems has been rapid since then. Such systems typically use one or more cameras mounted at an elevated position on the roadside to detect and track moving objects as well as detect traffic conflicts~\cite{meng2023roco}. To detect road objects, different methods have been used, to name a few, background subtraction \cite{furuya2014road}, frame difference \cite{messelodi2005computer}, synthesizing training data with AR and GAN~\cite{zhang2023robust}, feature-based detection \cite{saunier2006feature}, KanadeLucas-Tomasi tracking \cite{li2016robust},  cascading classifiers \cite{faisal2020automated} and many more \cite{datondji2016survey}. 

In recent years, with the development of deep
learning (DL) techniques in computer vision,
real-time high-quality vision-based perception algorithms
are proposed and widely applied into
ADAS systems as well as automated vehicles.
Though transformer-based object detectors
achieved state-of-the-art detection accuracy
~\cite{carion2020end,meng2021conditional,chen2022conditional,zhu2020deformable,liu2022dab,chen2022group}, the YOLO~\cite{redmon2016you,redmon2017yolo9000,redmon2018yolov3,bochkovskiy2020yolov4,ge2021yolox} series object detectors are known for
lightweight and fast processing
and suitable for CAV applications.
Vehicle trajectories are valuable data
for transportation research.
Visual tracking algorithms are designed
to obtain the object trajectories from
consecutive video frames.
There are two classes of object tracking.
single object tracking (SoT)~\cite{barina2016gabor,hariyono2014moving,kim2015moving,yao2020video,patel2013moving,taylor2021optimized} and
multiple object tracking (MoT)
~\cite{bae2014robust,berclaz2011multiple,bergmann2019tracking,dicle2013way,milan2013continuous,bewley2016simple}.
MoT methods can be deployed
in roadside perceptions for object trajectory
tracking.

These new DLs are also adopted in roadside vehicle perception and cooperative driving. For instance,  \cite{aboah2021vision} introduces a method that detects vehicles from roadside based on YOLOv5 detector and finds anomaly behavior using decision tree. \cite{huang2022rd} proposes a modified version of YOLOv5, that optimizes the performance for roadside perception tasks. In general, DL-based methodology for vehicle detection is at the initial stage, but of significant potential, thereby attracting increasing attention. 

In addition to the aforementioned systems using regular cameras, a wide range of other sensors have also been explored to overcome the limitations of regular cameras. For example, systems with other types of cameras are investigated, for instance, fish-eye cameras to enhance per-camera coverage \cite{wang2015real}, and thermal cameras to increase the robustness at night and in different weathers \cite{gaszczak2011real, iwasaki2013robust}. Recently, roadside lidars are explored in several researches for roadside vehicle detection \cite{sun2022object}. These methods include traditional point cloud detection methods based on background subtraction and point clustering \cite{zhang2018background, zhao2019detection, zhang2019vehicle}, as well as DL-based methods \cite{zhou2022leveraging, bai2022pillargrid}. \cite{wu2020vehicle, wu2020automatic} proposes lidar-based vehicle detection methods that are robust in adverse weather, especially in snow and heavy rain.

Furthermore, methods using multiple sensors and fusion strategies have been proposed for roadside perception. In \cite{shan2021demonstrations}, a roadside perception system with lidar and dual cameras is prototyped and demonstrated with CAVs self-driving using the roadside perception information only.  In \cite{zhang2022design, zou2022real}, multiple fish-eye cameras and thermal cameras are utilized for vehicle detection. \cite{du2021novel} provide a sensor fusion method that synchronizes detection results with millimeter-wave radar and camera detection. For some CAV applications, the ability
of predicting the objects' future positions
is also required.
Trajectory prediction methods often
use objects' historical trajectories
with semantic map information
to predict objects' future trajectories
~\cite{gao2020vectornet,cui2019multimodal,liang2020learning,wang2022ganet} or directly use raw sensor data to predict
future trajectories~\cite{luo2018fast,meng2021hymo}.

On the other hand,  V2X communications are proposed to transmit roadside perception results to CAVs in a real-time fashion.  \cite{rauch2011analysis}  provides a design using a V2X communication system for cooperative perception, and reports experimental results conducted by 802.11p radios and measures the transmission latency of cooperative awareness messages (CAM). Their research has found that the transmission latency of CAMs is within 20ms, but the overall latency can be as large as 200ms considering the idle time between periodical transmissions. In \cite{rauch2012car2x}, a vehicle-roadside cooperative perception system is proposed with a high-level fusion strategy. In \cite{tsukada2019cooperative}, the authors report the design of a communication system that broadcasts proxy-CAM to CVs and some initial field-tests results on communication reception rate and latency. The work is continued in \cite{tsukada2020autoc2x, tsukada2020networked}, a roadside perception unit is implemented with open-source autonomous driving software, inter-connected with RSU and sensors with high-speed networks. Thorough latency analysis is carried out using OMNET++ and field experiments.  \cite{yang2022scalable} systematically designs scalable communication network structures for infrastructure-vehicle cooperative driving tasks.


All aforementioned works concentrate on distinct aspects of a roadside perception system, without systematic integration, testing, and field implementation. In this work, we endeavor to orchestrate a comprehensive framework, encompassing object detection, localization, tracking, prediction, vehicle list encoding, message forwarding, and vehicle-side decoding. This integrative approach is a pivotal advancement towards achieving scalable, production-grade deployments in real-world scenarios. Moreover, the evaluation methods used in this paper focus on end-to-end perception performance for CAV applications, offering a more extensive exploration compared to existing works that primarily resolve singular issues or components. Consequently, this research serves not only as an academic advancement but also as a meticulously crafted template, paving the way for pragmatic deployments in the field of roadside perception.

\section{System Design}
\subsection{The Edge-Cloud System Design}
As mentioned above, MSight is a specialized, real-time roadside perception system, predominantly devised for CAVs. It bifurcates into two critical components: the roadside portion and the cloud portion. The former revolves around an edge device, serving as the nexus that orchestrates all roadside sensors, executes delay-sensitive applications, and fosters communication with CAVs through RSUs. Concurrently, the cloud component is engineered to be highly scalable, capable of assimilating expansive data streams. Data are archived in cloud storage and disseminated to various services via a publish/subscribe mechanism within the cloud. Together, these two portions construct an intricate edge-cloud system architecture, endowed with the versatility to accommodate a diverse array of downstream applications.

\begin{figure*}[ht]
    \centering
    \includegraphics[width=.95\linewidth]{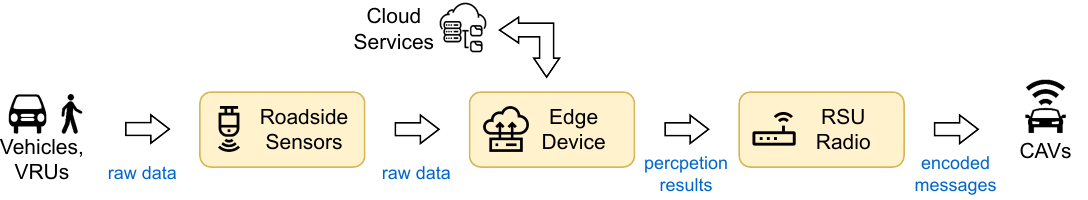}
    \caption{An illustration of the roadside portion of MSight system}
    \label{fig:roadside}
\end{figure*}

\subsection{The Roadside Portion}
\begin{figure}[hb]
     \centering
     \begin{subfigure}[b]{0.3\linewidth}
         \centering
         \includegraphics[width=\textwidth]{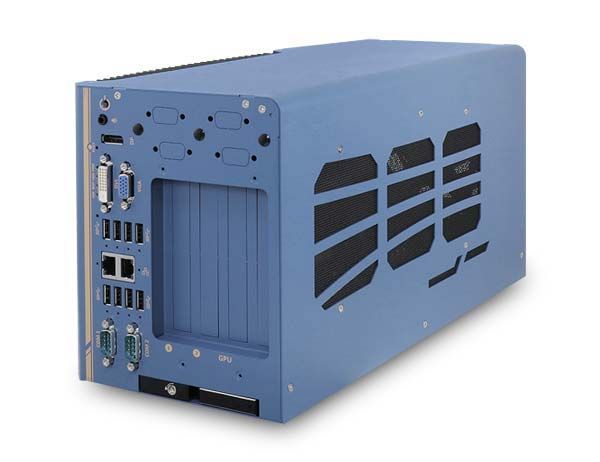}
         \caption{Nuvo edge device}
         \label{fig:nuvo}
     \end{subfigure}
     \begin{subfigure}[b]{0.3\linewidth}
         \centering
         \includegraphics[width=\textwidth]{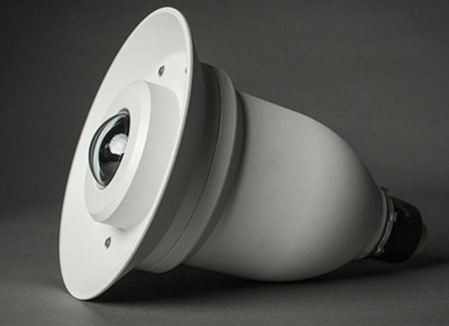}
         \caption{GridSmart fisheye camera}
         \label{fig:gs}
     \end{subfigure}
     \hfill
     \begin{subfigure}[b]{0.35\linewidth}
         \centering
         \includegraphics[width=\textwidth]{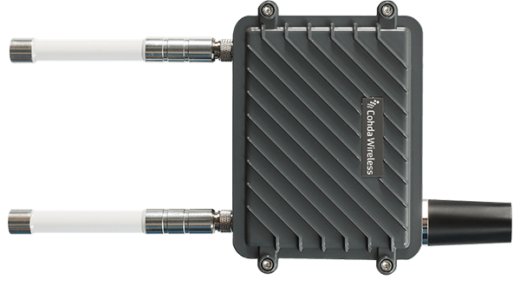}
         \caption{Cohda RSU MK5 radio}
         \label{fig:cohdaRSU}
     \end{subfigure}

        \caption{Nuvo edge device}
        \label{fig:devices}
\end{figure}
The principal objective of the roadside portion is to execute the perception algorithm, detailed in section \ref{section:algo}, and to convey the resulting perceptions to CAVs with minimal latency, as illustrated in Figure \ref{fig:roadside}. Vehicles and Vulnerable Road Users (VRUs) are identified through roadside sensors and the raw data they yield (such as raw images) are channeled to the edge device. This device processes the raw data using the perception algorithm and transmits the encoded perception results to CAVs via the RSU radio. This entire perception loop is localized to the roadside to guarantee minimal latency in the communication network. Simultaneously, the edge device relays selected data to the cloud, catering to the needs of other applications and services residing therein.

\begin{figure*}[t]
    \centering
    \includegraphics[width=.98\textwidth]{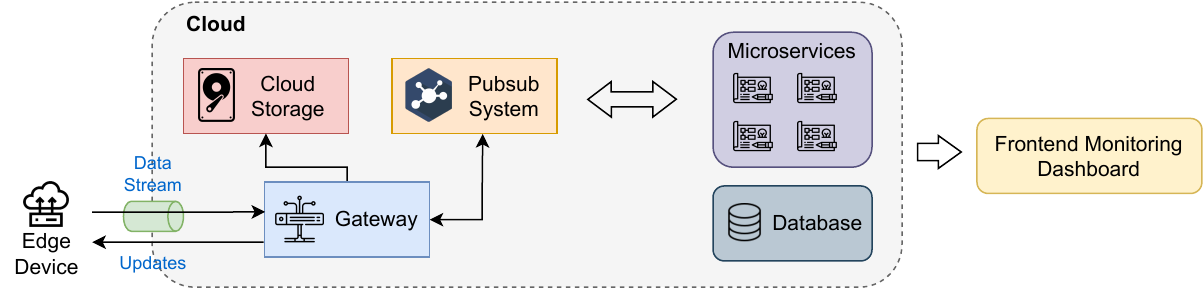}
    \caption{An illustration of the cloud-side portion of the MSight system}
    \label{fig:cloudside}
\end{figure*}

Figure \ref{fig:devices} displays the array of devices installed at the roadside. While MSight is engineered to accommodate a wide range of sensors and radios, the devices discussed in this paper pertain specifically to our installation described in section \ref{section:deployment}. Generally, the roadside component is subdivided into three primary elements: an edge device that operates as the nucleus, sensors that acquire raw data, and a radio that communicates with CAVs.
Figure \ref{fig:nuvo} illustrates the Nuvo edge device situated in the traffic cabinet near the roundabout, interconnected with sensors and RSU radio within the same network. GridSmart fisheye cameras are selected (as shown in Figure \ref{fig:gs}), for their superior field of view compared to conventional pin-hole cameras. These cameras are strategically positioned at each quadrant (north-east, north-west, south-east, south-west) of the roundabout.
Figure \ref{fig:cohdaRSU} presents the Cohda MK5 RSU radio, which is commissioned to periodically transmit messages encoded by the edge device.

\begin{figure}
\centering
\includegraphics[width=.8\linewidth]{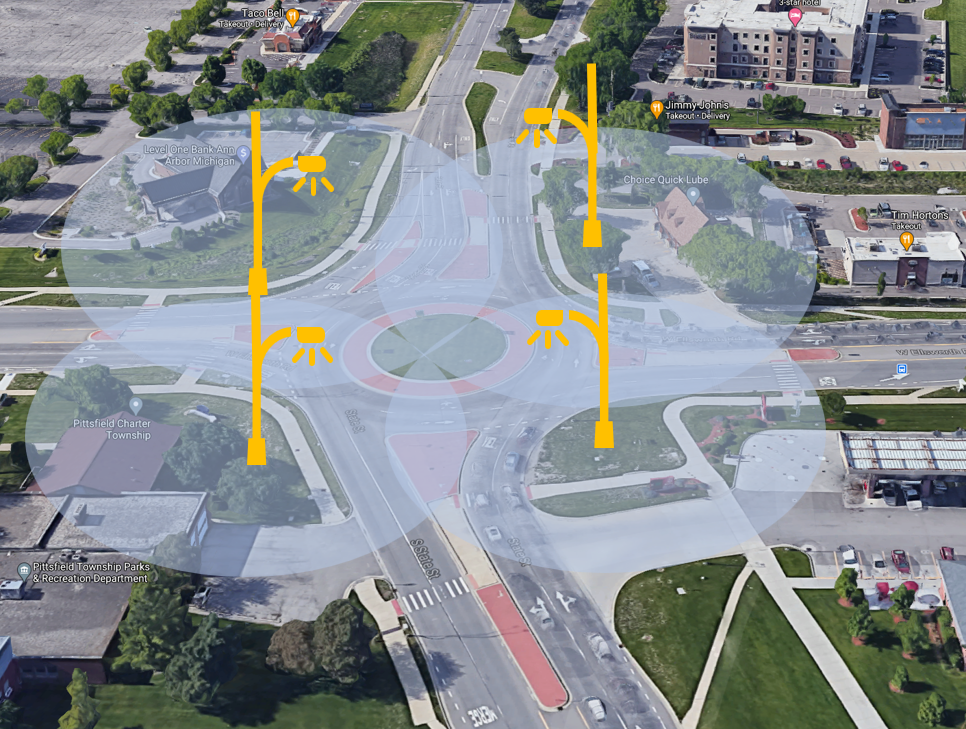}
\caption{The four corners of the roundabout installing the sensors, and the illustration of their coverage}
\label{fig:sensor-installation}
\end{figure}

\begin{figure*}[t]
    \centering
    \includegraphics[width=\textwidth]{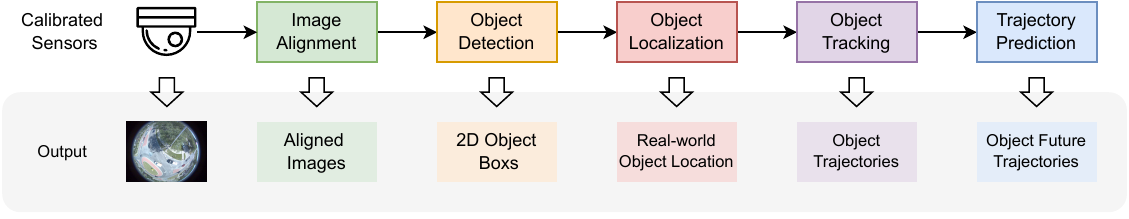}
    \caption{An illustration of overall proposed roadside perception framework}
    \label{fig:perception-arch}
\end{figure*}

\subsection{The Cloud Portion}

Figure \ref{fig:cloudside} illustrates the architectural design of MSight's cloud segment, constructed for extensibility and versatility. It incorporates a gateway to establish an API endpoint for the edge device, facilitating the streaming of roadside data. Once data reaches the gateway, it is relayed to a publication subscription (pub/sub) service, operating as a message dispatcher, distributing data to all its subscribers. This architecture ensures the decoupling of message producers and consumers, enhancing the system's robustness, scalability, and error resilience. New services can be seamlessly integrated to consume the data stream without altering existing cloud components. The pub/sub system to forward messages with minimal latency, enabling near real-time services. This feature is particularly beneficial for third parties requiring instantaneous traffic data.

Each application within the cloud is constructed as a microservice, ensuring scalability and ease of maintenance. These services, using the aforementioned pub/sub system for data communication, are completely decoupled, enabling the streamlined development of additional services and affirming MSight as a highly extendable system.

Moreover, a distinct dedicated link is established between the gateway and cloud storage, allowing the direct storage of data streamed from roadside edge devices. This link is deliberately isolated from the pub/sub system and all other microservices residing on the cloud, safeguarding data security by design. This structured and secure approach ensures the integrity and reliability of MSight's cloud infrastructure, accommodating a range of applications and services.

\subsection{Deployment}
\label{section:deployment}
This paper focuses on a particular deployment of MSight at an intersection between Ellsworth Road and State Street, Ann Arbor, Michigan, USA. Camera sensors are mounted on the light pole at the roadside of the roundabout. For this particular deployment, GridSmart fisheye cameras are selected due to their large coverage area. A Cohda RSU radio is also mounted on the light pole. There are four fisheye cameras mounted at the four corners of the roundabout, as delineated in Figure \ref{fig:sensor-installation}. Additionally, a provisional coverage area for each sensor is represented in the figure, with each sensor monitoring a quarter of the roundabout.

\section{Perception Algorithm}
\label{section:algo}

The perception algorithm aims at detecting, tracking, and predicting
objects such as cars, trucks, buses, and motorcycles at the roundabout
in real-time.
The proposed framework consists of 6 components:
camera calibration,
image alignment, object detection, 
object localization,
object tracking, and future
trajectory prediction. 
As shown in Figure~\ref{fig:perception-arch},
the components are assembled in a sequential manner.
Firstly, we calibrate the sensors
to get the homography transformation matrices
of sensors for image pixel location to
real-world coordinate projection.
Since the roadside camera view might slightly change
over time, the calibration might be inaccurate.
We align the input image to a standard
camera view to guarantee the calibration
accuracy.
A YOLOX~\cite{ge2021yolox}-based lightweight object detector is applied for
$2$D object detection.
For object localization,
we first project the object pixel locations
to real-world coordinates,
then we fuse the object coordinates
from multiple cameras together
to form a unified object location result.
A SORT~\cite{bewley2016simple}-based object tracking
algorithm is used to extract object
trajectories.
With the historical trajectories,
a transformer-based trajectory prediction
module is applied to predict the object's future
trajectories.

\subsection{Camera Calibration}

\begin{figure*}[ht]
    \centering
    \includegraphics[width=.99\linewidth]{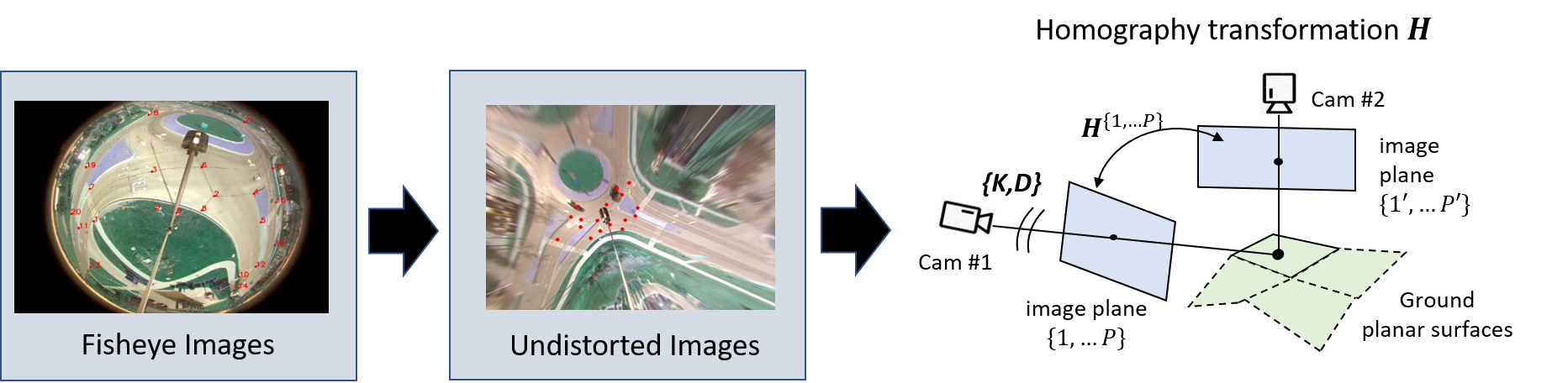}
    \caption{The landmark-based calibration method for fisheye camera}
    \label{fig:calibration}
\end{figure*}

Traditional checkerboard calibration methods are inefficient in this case, as they require field operations, specialized devices, and skilled operators on site. Therefore, we use a landmark-based calibration method. Figure \ref{fig:calibration} shows a set of landmark points selected for the calibration. Landmarks are identified both in the fisheye images 
produced by our fisheye camera and in satellite images in which the latitude and longitude of the points are known.

A calibration method is then developed using these landmark pairs from the two images. Figure \ref{fig:calibration} illustrates the calibration method. Assume that the camera lens follows a generic radially symmetric model. $r(\theta)=k_1\theta + k_2\theta^3 + k_3\theta^5 + \dots$ where $\theta$ is the angle between the principle axis and the incoming ray,  $r(\theta)$ is the distance between the correspondence point to the principle point \cite{kannala2006generic}. An estimation of the intrinsic matrix of the camera and its distortion coefficients can be made according to \cite{shah1996intrinsic}. Then, a homography transformation can be found between the undistorted image and the world latitude and longitude with least square regression and RANSAC consensus between the two groups of landmark sets.  This transformation can be applied to map any point in the fisheye image to the world latitude and longitude.

\begin{figure}[t]
    \centering
    (a)\includegraphics[angle=90,width=.2\textwidth]{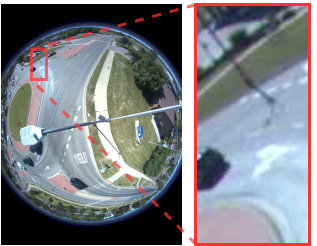}
    (b)\includegraphics[angle=90,width=.2\textwidth]{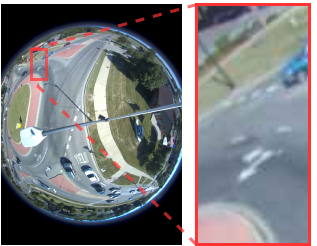}
    \caption{An illustration of camera view change within the
    same day at (a) 10am and (b) 6pm}
    \label{fig:camera-view-change}
\end{figure}

\begin{figure}[t]
    \centering
    \includegraphics[width=.45\textwidth]{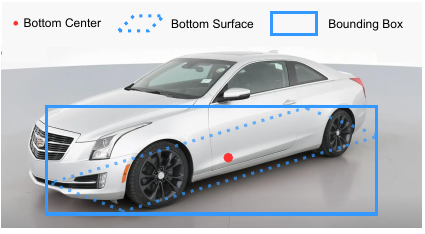}
    \caption{An illustration of the bounding box annotation.
    We use the minimum bounding rectangle of object bottom surface
    as object bounding box for object detection.}
    \label{fig:bbox-annotation}
\end{figure}

\subsection{Image Alignment}
The roadside cameras are installed on four poles
at four corners of the roundabout.
Since the roadside cameras will serve
for an extended period of time once installed,
both elastic deformation of the poles caused by temperature change,
or metal creep of the poles could happen.
This will lead to a change of camera view.
As shown in Figure~\ref{fig:camera-view-change},
The camera view at 10am and 6pm are different
even on the same day.
In the zoomed-in region, one can see that the
arrow landmark at 6pm is higher than at 10am.
These view changes, if not compensated,
will lead to calibration errors.

To compensate for the view changes, we apply
a run-time image alignment module.
The module contains two parts:
geometric transformation estimation
and image wrapping.  
For geometric transformation estimation,
given standard image $\mathbf{I}_s$
and input image $\mathbf{I}_i$,
we first convert them to grayscale images
$\hat{\mathbf{I}}_s$ and $\hat{\mathbf{I}}_i$.
Then we use Enhanced Correlation Coefficient~\cite{evangelidis2008parametric}
to compute the geometric transformation
$\mathbf{T}$ between two images.

\begin{align}
    \mathbf{T} = \arg\max_{\mathbf{W}}
    \operatorname{ECC}(\hat{\mathbf{I}}_i(x, y), \hat{\mathbf{I}}_s(x', y'))
\end{align}
Here
\begin{align}
    \begin{bmatrix}
        x' \\
        y'
    \end{bmatrix}
    = \mathbf{W}
    \begin{bmatrix}
        x \\
        y \\
        1
    \end{bmatrix}
\end{align}

With the geometric transformation $\mathbf{T}$,
we can warp the input image to the same view as
the standard image:
\begin{align}
    \tilde{\mathbf{I}_i} = \operatorname{WarpPerspective}(\mathbf{I}_i, \mathbf{T})
\end{align}

Then we feed the wrapped image $\tilde{\mathbf{I}_i}$
into object detector for object detection.

\subsection{Object Detection}
For object detection, we designed three object categories:
car, truck/bus/trailer, and motorcyclist/cyclist/pedestrian.
For object bounding boxes, instead of using the object
2d bounding boxes, we use the minimum bounding rectangle (MBR)
of objects' bottom surface as illustrated in Figure~\ref{fig:bbox-annotation}.
In this way, the center point of the bounding box will be
the bottom center of the object.

\begin{figure}[t]
    \centering
    \includegraphics[width=.49\textwidth]{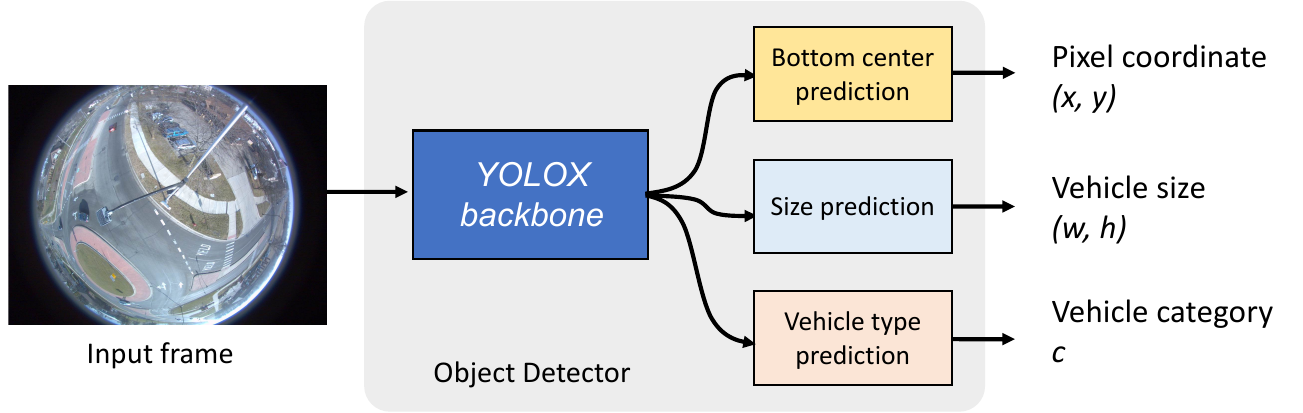}
    \caption{An illustration of the object detection
    pipeline}
    \label{fig:object-detection}
\end{figure}

\paragraph{Network}
A standard {YOLOX-nano}~\cite{ge2021yolox} model is used for object detection.
The {YOLOX-nano} is a lightweight single-stage object detector
that contains $0.91$M parameters and $2.56$G FLOPs with input image
size $640\times 640$.
The {YOLOX-nano} model contains a DarkNet backbone,
a feature pyramid network (FPN) layer, and YOLOX detection head.
The backbone extract deep feature from the input image,
and the FPN layer aggregates multi-scale features
output by the backbone.
The multi-scale features are fed into the YOLOX detection
head for classification and bounding box regression.
Fig~\ref{fig:object-detection} shows an illustration
of the object detection pipeline.

\paragraph{Training}
For the object detector training,
we use $5,000$ images in total as the training
set, and the four camera views share the
the same network.
For data augmentation,
the translation, sheer, rotation,
and hsv augmentations are adopted.
The model is trained for $150$ epochs
with an initial learning rate $5e-5$.
The learning rate is dropped by a
factor of $10$ at epoch $100$.
The mini-batch size used for training
is $8$.

\subsection{Object Localization}
Once we obtained the object $2$D location
in the image, we need to 
obtain the real-world object coordinates
for down-stream applications.
In the object localization module,
we first map the object $2$D location
to real-world latitude and longitude
coordinates, then we fuse
the object location information
for multiple cameras together
to get a merged object real-world
locations across different camera views.
\paragraph{Image to real-world mapping}
For image position $(x_p, y_p)$,
we first obtain its undistorted position
$(\hat{x}_p, \hat{y}_p)$ with the
intrinsic parameter $K$ and
distortion parameter $D$ of the fisheye camera.
Then we apply the homography transformation
$\mathbf{H}$
obtained by camera calibration
to the undistorted image to get
the real-world coordinate $(x_r, y_r)$.

\begin{figure*}[t]
    \centering
    \includegraphics[width=.95\textwidth]{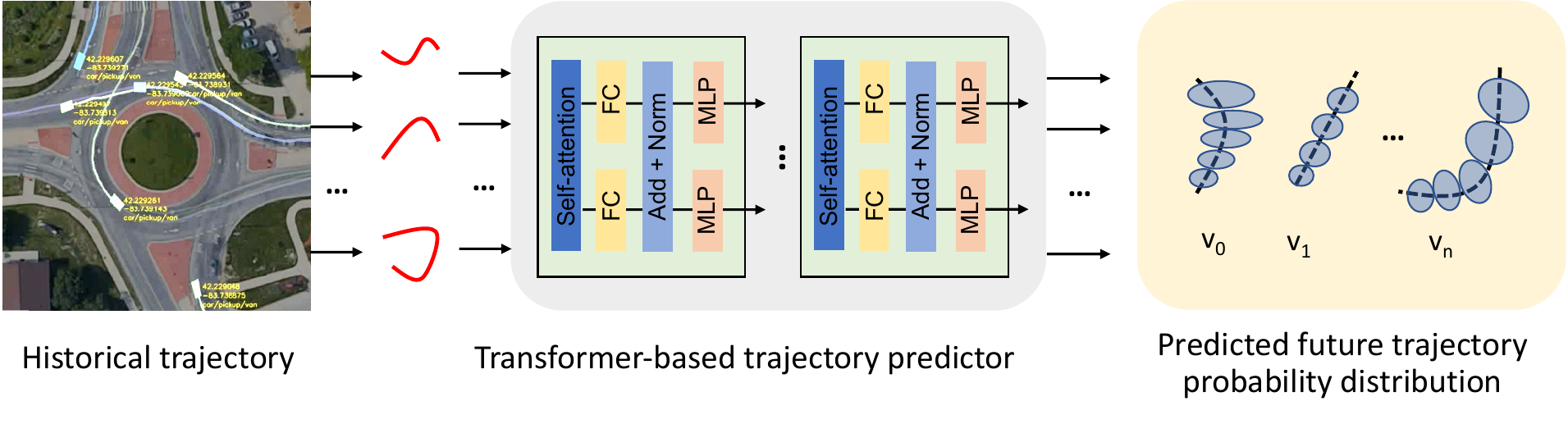}
    \caption{An illustration of the object prediction
    pipeline}
    \label{fig:object-prediction}
\end{figure*}

\paragraph{Multi-camera Fusion}
Since we installed four fisheye cameras
at the four corners (Northeast, Northwest,
Southeast, Southwest) at the roundabout,
the detection results of the four cameras
needs to be fused together
to obtain a merged detection result of the whole
roundabout.
We split the roundabout into four parts
(NE, NW, SE, SW)
corresponding to the four camera locations.
We assign each part as the region of interest
of each camera.
We only select the detected objects within
the region of interest of each camera
and put the detected objects together
to obtain the fused detection results.

\subsection{Object Tracking}
The object tracking module is built based
on SORT~\cite{bewley2016simple}, an online object tracking algorithm.
The tracking module consists
a Kalman Filter~\cite{welch1995introduction}
for state estimation
and a Hungarian Algorithm~\cite{kuhn1955hungarian}
for the association
between existing targets
and detections.
Compared to the SORT algorithm,
we use the predicted
future locations of each target
provided by our transformer-based
trajectory prediction module instead
of the Kalman Filter
for association for better
tracking performance.
Following SORT,
the state of each target is defined as
\begin{align}
    \mathbf{s} = [x_c, y_c, s, r, v_x, v_y, v_s, v_r]^\top
\end{align}
Here, $x_c$ and $y_c$ are the real-world
latitude and longitude coordinate
of the target,
$s$ and $r$ are the scale and the
aspect ratio of the target's bounding box
respectively.
$v_x, v_y, v_s, v_r$ are the derivatives
of $x, y, s, r$.

To assign detections to existing targets,
we first compute the intersection-over-union (IoU)
between detections and all predicted boxes
of each target.
Then we use the IoU matrix as the
cost matrix of the Hungarian Algorithm
for bipartite matching.
If a target is not detected in $3$
consecutive frames, we will delete the target
from the scene.

\begin{figure}[h]
    \centering
    \includegraphics[width=.48\textwidth]{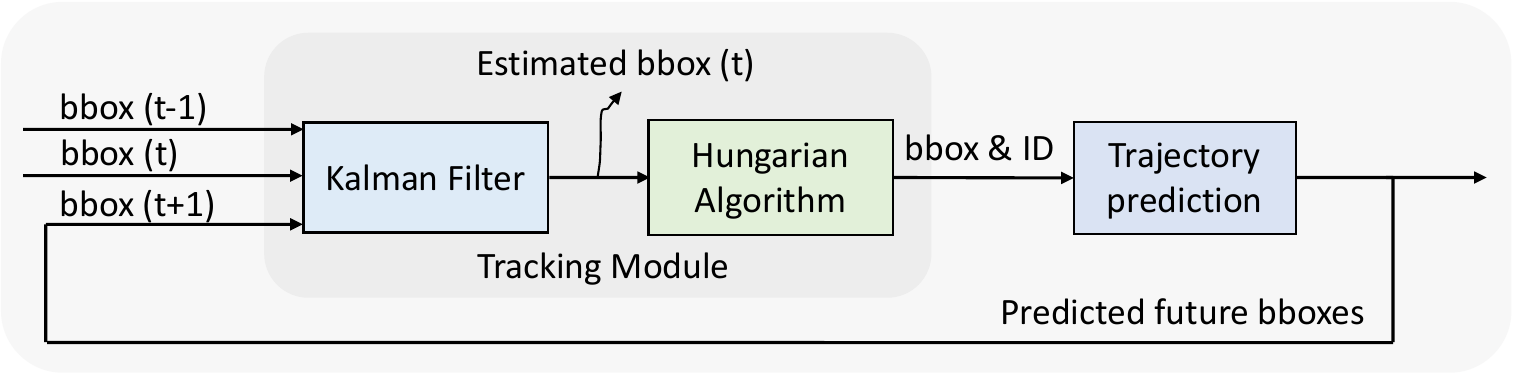}
    \caption{An illustration of the object tracking
    pipeline}
    \label{fig:object-tracking}
\end{figure}

\subsection{Future Trajectory Prediction}
Trajectory prediction aims at predicting
the future locations of objects.
For an object,
With the historical object positions
obtained by the object tracking module
\begin{align}
    \mathbf{x}^{(t)} &= ({x_c}^{(t)}, {y_c}^{(t)}) \\
    \mathbf{X} &= (\mathbf{x}^{(t-5)}, 
    \mathbf{x}^{(t-4)}, ...,
    \mathbf{x}^{(t)})
\end{align} 
We feed the historical positions
of all objects $\mathbf{X}_1, \mathbf{X}_1, ...,
\mathbf{X}_n$
into a transformer $t$~\cite{vaswani2017attention},
and predict the future positions of the objects.
As shown in Fig~\ref{fig:object-prediction},
the transformer consists of multiple 
transformer encoder layers.
Each transformer encoder layer contains 
a self-attention layer,
a fully-connected (FC) layer,
an Add + Norm layer,
and a multi-layer perception (MLP) layer.

\paragraph{Input encoding}
With the input historical trajectories of objects
$\mathbf{X}_1, \mathbf{X}_2, ..., \mathbf{X}_n$,
first we apply a positional mapping~\cite{vaswani2017attention,carion2020end}
to obtain a higher dimensional vector
and enable the prediction to easily
approximate a higher frequency function.
for each object historical trajectory
$\mathbf{X}$, the positional mapping $\mathcal{T}$
is
\begin{align}
    \mathcal{T}: \mathbf{X} \mapsto (\mathbf{X},
    \sin{2^0\pi \mathbf{X}}, \cos{2^L\pi \mathbf{X}}, ...,
    \sin{2^0\pi \mathbf{X}}, \cos{2^L\pi \mathbf{X}})
\end{align}
$L$ is the positional mapping length parameter of $\mathcal{T}$.
The output vector $\mathcal{T}\mathbf{X}$ is a vector with
length $2L + 1$.

\paragraph{Transformer-based trajectory predictor}
The trajectory predictor contains
a linear layer to increase the dimension of the vectors
to $256$,
a transformer with $4$ encoder layers,
and a predictor to predict trajectory mean and variance.
In each encoder layer, there is a multi-head
self-attention layer with the head number $8$.
The self-attention allows each object trajectory
to interact with other object trajectories.
This enables the predictor to predict
complex driving behaviors like yielding
to other vehicles.
The MLP layer allows interactions within each object
across different frames.
The MLP layer contains a FC layer,
a GeLU layer~\cite{hendrycks2016gaussian} and another FC layer.
For the prediction head,
two linear layers are used to predict
future object location mean and variance
accordingly.

\paragraph{Output encoding}
The trajectory predictor predicts
the object positions in the future $5$ frames.
The interval between two frames is $0.4$ seconds.
For each frame, we model the possible
object position distribution as a Gaussian
distribution $\mathcal{N}(\mu, \sigma^2)$.
The mean $\mu$ and variance $\sigma^2$
are predicted.

\paragraph{Loss function}
We use regression loss with uncertainty
estimation as the training loss of the
trajectory predictor.
For each future frame and each object,
we denote the object future position
as $(x, y)$, the
predicted object position mean
as $(\mu_x, \mu_y)$,
and predicted object position variance
as $(\sigma_x, \sigma_y)$.
The losses for the predictor training are
\begin{align}
    \ell(\mu) &= (x - \mu_x)^2 + (y - \mu_y)^2 \\
    \ell(\sigma_x^2) &= ((x - \mu_x)^2 - \sigma_x^2)^2 \\
    \ell(\sigma_y^2) &= ((y - \mu_y)^2 - \sigma_y^2)^2 \\
    \mathcal{L} &= \ell(\mu) + \ell(\sigma_x^2) + \ell(\sigma_y^2)
\end{align}

\section{Experimental Evaluations}
\subsection{Experiment setup}

\begin{figure*}
    \centering
    \includegraphics[width=.23\textwidth, height=.23\textwidth]{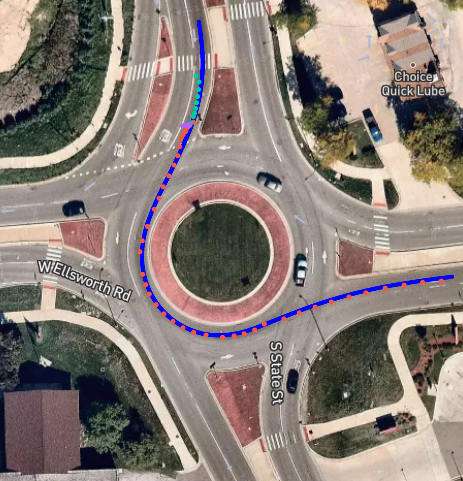}~~
    \includegraphics[width=.23\textwidth, height=.23\textwidth]{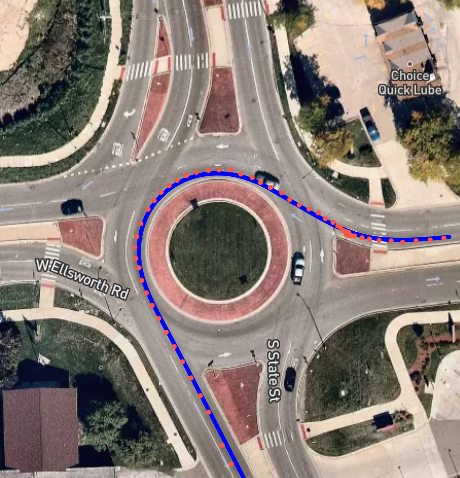}~~
    \includegraphics[width=.23\textwidth, height=.23\textwidth]{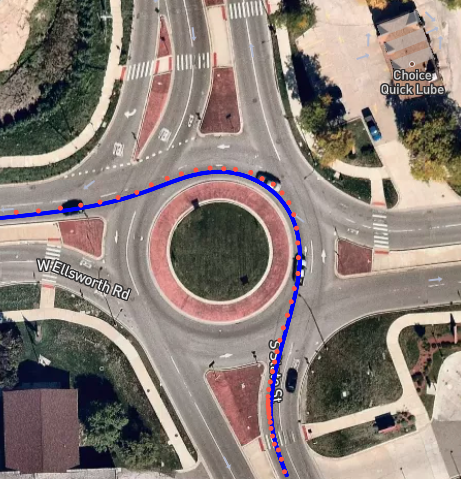}~~
    \includegraphics[width=.23\textwidth, height=.23\textwidth]{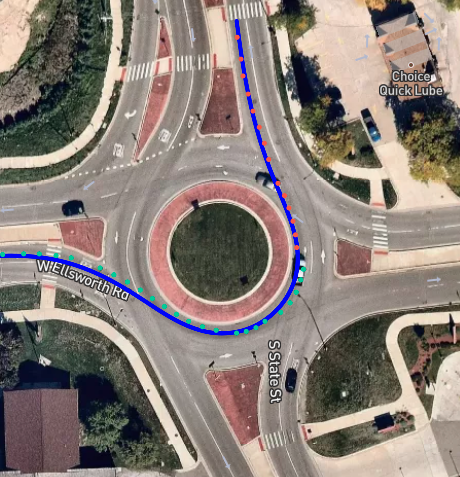} \\
    Trip $1$\hspace{3.7cm}Trip $2$\hspace{3.7cm}Trip $3$\hspace{3.7cm}Trip $4$ \\
    \vspace{2mm}
    \includegraphics[width=.23\textwidth, height=.23\textwidth]{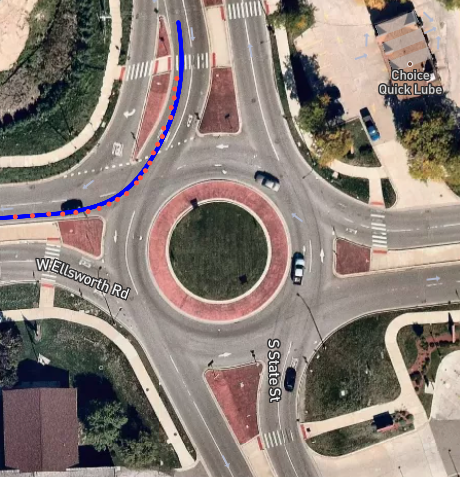}~~
    \includegraphics[width=.23\textwidth, height=.23\textwidth]{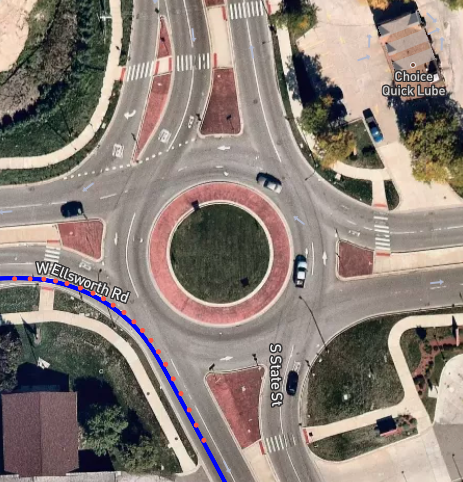}~~
    \includegraphics[width=.23\textwidth, height=.23\textwidth]{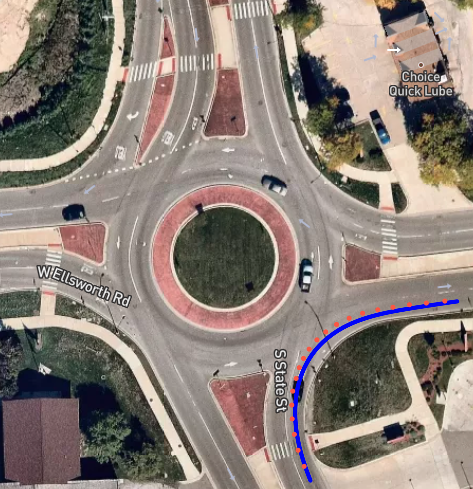}~~
    \includegraphics[width=.23\textwidth, height=.23\textwidth]{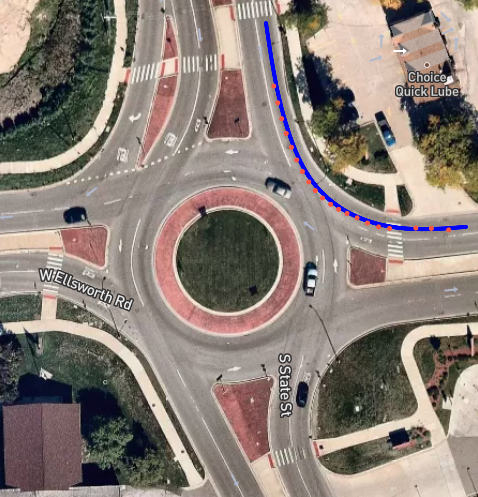} \\
    Trip $5$\hspace{3.7cm}Trip $6$\hspace{3.7cm}Trip $7$\hspace{3.7cm}Trip $8$ \\
    \caption{Visualizations of the $8$ trips in the field-test.
    The blue lines are the vehicle's ground-truth
    trajectories obtained by RTK.
    The colored dots are the detected vehicle locations
    obtained by MSight perception system with $4$ cameras setup. Different
    colors mean different tracking IDs.}
    \label{fig:trips}
\end{figure*}

\vspace{1mm}
\noindent\textbf{Training dataset.}
The training dataset contains $4,000$ images,
$1,000$ images for each camera (NE, NW, SE, SW).
The images are collected in May and June of 2021.
There are three object categories in the training
dataset: cars, trucks/trailers/buses, and other road users.

\vspace{1mm}
\noindent\textbf{Training setup.}
For object detection, we use the training
pipeline of YOLOX~\cite{ge2021yolox}.
We train the YOLOX-nano model for $150$ epochs
with a mini-batch size $8$.The initial learning
rate is $5e-5$ and decayed with a factor of
$10$ after $100$ epochs. The weight decay is
set to be $5e-4$ and Adam~\cite{kingma2014adam} optimizer
is used. The model is trained on PyTorch~\cite{paszke2019pytorch}
$1.9$ platform with a NVIDIA RTX $3080$ GPU.
The input image size is $640\times 640$.
The data augmentation we used in training is the
same as YOLOX's:
random resize, horizontal flip, rotation,
translation, shearing, and color distortion
are applied.

\vspace{1mm}
\noindent\textbf{Evaluation setup.}
\begin{figure}
    \centering
    \includegraphics[width=\linewidth]{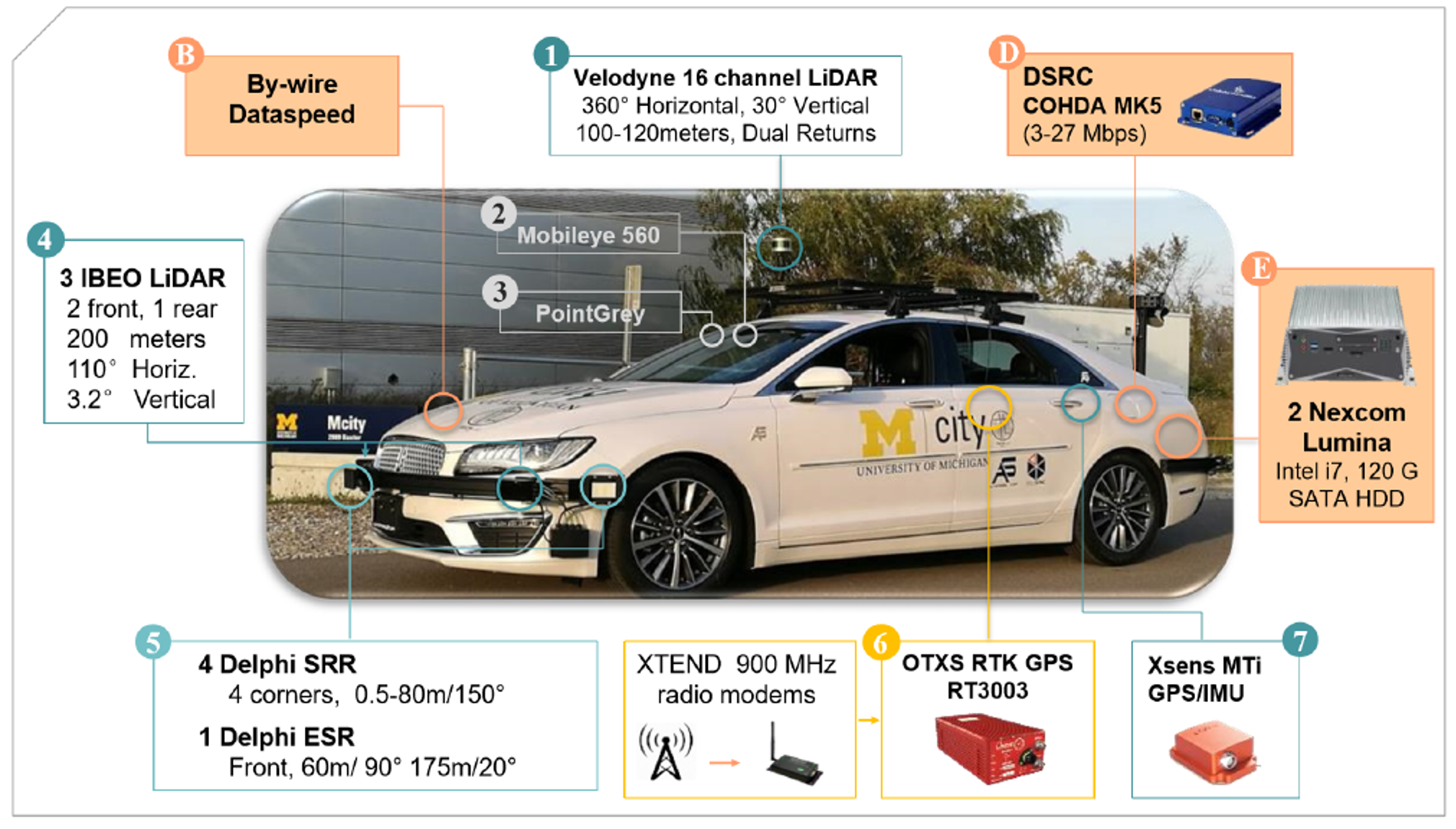}
    \caption{An illustration of the vehicle
    for the field test. The RTK GPS, IMU, and V2X module
    are used for perception and system latency evaluations.}
    \label{fig:mkz}
\end{figure}
We evaluate our infrastructure-based perception system
at the same roundabout where we installed the system.
The evaluation experiment proceeded on a Friday afternoon
in October 2022 with a heavy traffic volume.
A connected automated vehicle\footnote{\url{https://mcity.umich.edu/}}
-- a Hybrid Lincoln MKZ~\ref{fig:mkz}, which is equipped
with a high-precision RTK unit and an IMU is used
for evaluation. The evaluation contains $8$ trips at the roundabout:
$4$ trips contain a $270$-degree turn 
using the inner lane, $4$ trips contain a $90$-degree
turn using the outer lane.
The routes of the trips are illustrated in Figure~\ref{fig:trips}.
The evaluation area is set to be the circular area
around the roundabout within a $50$ m radius.
The sampling frequency
of the RTK is $50$ Hz, and the RTK latency is negligible.
We use the vehicle
trajectory produced by RTK as ground truth,
and compare it with the trajectory detected
from the perception system.

\subsection{Perception evaluation metrics}
In the following, we describe the performance metrics of the MSight perception system. 

\vspace{1mm}
\noindent\textbf{Detection metrics.}
For object detection, use standard
evaluation metrics including False Negative rate,
False Positive rate, lateral position error,
and longitudinal position error.
Detections with lateral position errors smaller
than $1.5$ m are regarded as True Positives,
otherwise False Positives.
The False Positive rate is defined as
\begin{align}
    \operatorname{FP\ rate} = \frac{|
    \operatorname{FP}|}{|
    \operatorname{Dets}|}
\end{align}
Here $|\operatorname{Dets}|$ means the 
total number of detections.
Given the detection frequency
(in our case, $2.5$ Hz),
the expected timestamps and locations of the detections
can be obtained. For each expected timestamp,
if the detection at the timestamp is missing,
or the detection lateral position error is larger
than $1.5$ m, a False Negative at this timestamp will
be collected.
\begin{align}
    \operatorname{FN\ rate} = \frac{|
    \operatorname{FN}|}{|
    \operatorname{gtDet}|}
\end{align}
Here $|\operatorname{gtDets}|$ means the expected
number of
ground-truth detections of the trip.
For the lateral position error, we
report the mean lateral position error of
the True Positive detections. 

\begin{figure}[t]
    \centering
    \includegraphics[width=.8\linewidth]{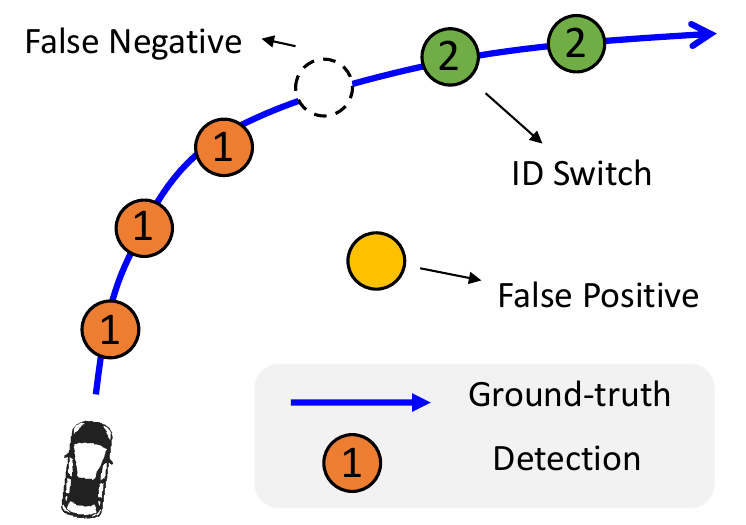}
    \hspace{2mm}
    \caption{An illustration of detection and tracking errors,
    including False Negative, False Positive, and ID Switch.}
    \label{fig:metrics}
\end{figure}

\vspace{1mm}
\noindent\textbf{Tracking metrics.}
For object tracking,
we report the number of ID switches (\#ID Switch),
longest track preservation time (Longest Track ($\%$)),
and Multi-Object Tracking Accuracy (MOTA).
The \#ID Switch means how many times the tracking ID
switches to another one for the same object.
For example, in Figure~\ref{fig:metrics},
at first the object is tracked with ID $1$,
then ID switched to $2$.
In this case, the \#ID Switch is $1$.
The longest track preservation time means
the number of frames
of the longest track divided by the number
of frames of the whole trip.
In Figure~\ref{fig:metrics},
the Longest Track is $50\%$.

MOTA is one of the most representative measures
of multi-object tracking.
MOTA measures three types of errors
mentioned above:
False Positives, False Negatives, and \#ID
Switch.
The MOTA is defined as
\begin{align}
    \operatorname{MOTA} = 1 - \frac{|\operatorname{FN}| + 
    |\operatorname{FP}| + |\operatorname{ID Switch}|}{|\operatorname{gtDets}|}
\end{align}
For perfect detection and tracking, the MOTA
score should be $1$. The larger the MOTA score
is, the fewer errors the model makes.

\vspace{1mm}
\noindent\textbf{Trajectory prediction metrics.}
For trajectory prediction, we use standard
Final Displacement Error (FDE) metrics.
Suppose we want to predict the object
location in next $K$ frames.
The predicted object locations are
$(\mathbf{x}_1, ..., \mathbf{x}_K)$,
and the ground-truth object locations are
$(\hat{\mathbf{x}}_1, ..., \hat{\mathbf{x}}_K)$.
The FDE is defined as
\begin{align}
    \operatorname{FDE}_K(m) = m(\mathbf{x}_K - \hat{\mathbf{x}}_K)
\end{align}
FDE measures the displacement
between the predicted location and ground-truth location
at the final frame.
Here $m$ is an error measure.
We use latitude position error
and longitudinal error as the error measures.
In our experiments, we set $K=3$.
With an interval of $0.4$ seconds between frames,
we predict the object locations in
the future $1.2$ seconds.

\subsection{Camera calibration and image alignment}
\noindent\textbf{Camera calibration.}
For each camera, we label
about $20$ landmarks both on the GoogleMap satellite image
and the camera-captured image.
Table~\ref{tab:calibration} shows the average
calibration error of the labeled landmarks.
For all four cameras, the calibration errors
are controlled within $0.6$ m, which ensures
the localization accuracy of our perception system.

\begin{table}[t]
    \renewcommand{\arraystretch}{1.3}
    \setlength{\tabcolsep}{8pt}
    \centering
    \begin{tabular}{l|c|c|c|c}
    \shline
        Camera & NE & NW & SE & SW \\
        \shline
        Calibration error & $0.53$ m & $0.47$ m & $0.39$ m & $0.46$ m \\
        \shline
    \end{tabular}
    \caption{Calibration errors of four cameras.}
    \label{tab:calibration}
\end{table}
\vspace{1mm}
\noindent\textbf{Image alignment.}
As mentioned, roadside cameras will serve for
a long time in their life cycle.
The pose of the cameras might change slightly
due to pole deformation.
The pose changes will lead to inaccurate calibrations
at different dates and times.
To compensate for the issue, we develop an image alignment
algorithm to align the images to a set of 
standard images.
This will preserve consistent calibration quality across
dates and times.
As shown in Table~\ref{tab:alignment},
we provide the calibration error at different dates and times.
The calibration is done at $10$ am, June $23$th, $2022$,
and we re-evaluate the calibration quality at $12$ pm, $2$ pm, $4$ pm, and $6$ pm of the same day,
as well as the same time but on July $23$th, $2022$, and 
August $23$th, $2022$.
A change of date or time can lead to a larger calibration error.
While, with image alignment, the calibration
error is controlled to around $0.46$ m,
which is the original calibration error without
change of date or time.

\begin{table}[t]
    \renewcommand{\arraystretch}{1.3}
    \setlength{\tabcolsep}{9pt}
    \centering
    \begin{tabular}{l|l|c|c}
    \shline
        Date & Time & Error (w/o align) & Error (w/ align) \\
        \shline
        June $23$th, $2022$ & $10$ am & $0.46$ m & $0.46$ m \\
        \hline
        June $23$th, $2022$ & $12$ am & $0.57$ m & $0.46$ m \\
        June $23$th, $2022$ & $2$ pm & $0.62$ m & $0.46$ m \\
        June $23$th, $2022$ & $4$ pm & $0.63$ m & $0.46$ m \\
        June $23$th, $2022$ & $6$ pm & $0.65$ m & $0.47$ m \\
        \hline
        July $23$th, $2022$ & $10$ am & $0.54$ m & $0.46$ m \\
        August $23$th, $2022$ & $10$ am & $2.97$ m & $0.46$ m \\
    \shline
    \end{tabular}
    \caption{Evaluations on the camera calibration and image alignment.
    The calibration is performed on June $23$th at $10$ am.
    This calibration has a larger error
    in other dates or times when without image alignment.
    With image alignment, the calibration errors are consistently
    small across different dates and times.}
    \label{tab:alignment}
\end{table}

\begin{table*}[t]
    \renewcommand{\arraystretch}{1.3}
    \setlength{\tabcolsep}{4.5pt}
    \centering
    \begin{tabular}{l|c|c|c|c|c|c|c|c}
    \shline
        \multirow{2}{*}{Method} & \multirow{2}{*}{Camera Setup}
        & \multicolumn{4}{c|}{Detection Evaluations}
        & \multicolumn{3}{c}{Tracking Evaluations} \\
        \cline{3-9}
        & & FN rate$\downarrow$ ($\%$)
        & FP rate$\downarrow$ ($\%$) & Lat. Error$\downarrow$ (m) & Lon. Error$\downarrow$ (m) & 
        \#ID Switch$\downarrow$ & Longest Track$\uparrow$ ($\%$) & MOTA$\uparrow$ \\
        \shline
        MSight & NE + NW + SE + SW & $11.83$ & $4.51$ & 
        $0.63$ & $0.90$ & $0.38$ & $90.25$ & $0.82$ \\
        MSight & NE + SW & $22.26$ & $9.69$ & 
        $0.63$ & $1.03$ & $0.00$ & $100.00$ & $0.66$ \\
        MSight & NE & $37.04$ & $30.48$ & 
        $0.74$ & $1.13$ & $0.13$ & $97.50$ & $0.42$ \\
        \shline
    \end{tabular}
    \caption{Evaluations of the MSight perception system.
    Three camera setups are tested: all four cameras,
    two cameras in diagonal directions, and only
    one camera.
    The camera setup with four cameras achieves 
    overall the best performance.
    The reported results are the mean results of $8$
    trips.}
    \label{tab:main-result}
\end{table*}

\subsection{Object detection and tracking results}
Figure~\ref{fig:trips} and
Table~\ref{tab:main-result}
show the perception results of our field test.
In Figure~\ref{fig:trips}, 
for all $8$ trips, the localization
accuracy of the MSight perception system
is satisfactory. The colored dots (detected points)
are overlapped with or
very near to the blue lines (ground truth).
For trip $1$ and trip $4$, there is one ID
switch. We check the raw videos for
these two trips, and find out that,
the testing vehicle is occluded by a truck
in the outer lane of the roundabout,
and causes the miss detection and ID switch.
We believe that this issue can be
alleviated when a more advanced tracking
algorithm is used.
Further, the coverage of the MSight perception
system is generally good.
The blue lines are within the region-of-interest
(within $50$ m radius of the roundabout center).
As shown in Figure~\ref{fig:trips},
MSight has perfect horizontal coverage,
while in the vertical direction,
MSight has small blind spots 
at the top and bottom areas.

\begin{figure}[t]
    \centering
    \includegraphics[width=.3\linewidth,
    height=.3\linewidth]{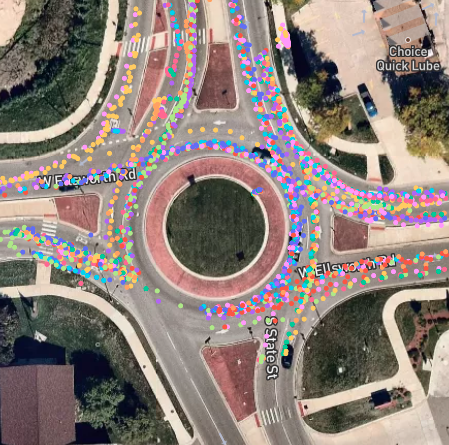}
    ~~\includegraphics[width=.3\linewidth,
    height=.3\linewidth]{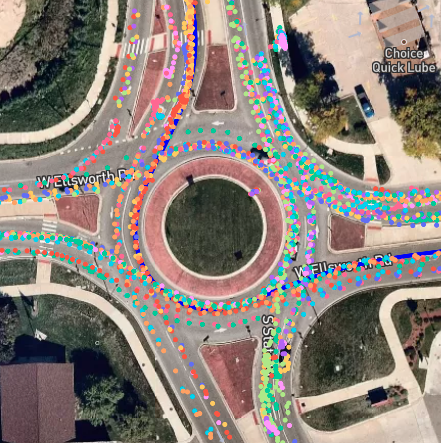}
    ~~\includegraphics[width=.3\linewidth,
    height=.3\linewidth]{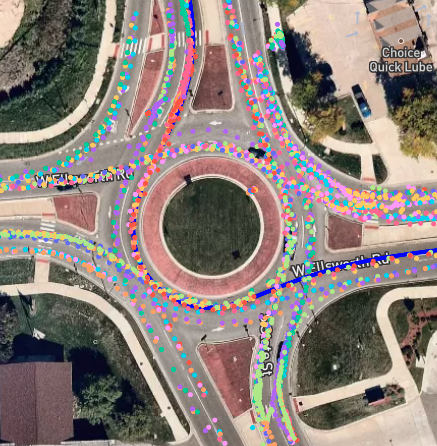} \\
    ~~~NE only~~~~~~~~~~~~~~NE + SW~~~~~~~~~~~four-camera
    \caption{Visualizations of detection point distributions
    of different camera settings. NE only setting is not able
    to cover the whole roundabout. NE + SW and four-camera
    setting both have good coverage.}
    \label{fig:detection-distribution}
\end{figure}

In Table~\ref{tab:main-result},
we show the quantitative evaluation results
of MSight perception system with different
camera setups.
There are three setups we used:
(1) all four cameras at the northeast, northwest,
southeast, southwest corner of the roundabout,
(2) only two cameras in the diagonal direction:
northeast and southwest corners,
and (3) only one camera at the northeast corner.

For the four-camera setup, the MSight has the best
detection quality with FN rate $11.83\%$,
FP rate $4.51\%$ and lateral error $0.63$ m.
However, in the four-camera setup, the MSight
has a higher \#ID Switch and shorter Longest Track.
The reason might be: more cameras cause 
more frequent switches
between cameras, which might lead to more \#ID Switches.
MSight with four-camera achieves the best MOTA score $0.82$.
For the two-camera (NE+SW) setup, the detection quality is worse
than the four-camera setup. The reason might be:
with only two cameras, each camera needs to cover a larger
area compared to the four-camera setup,
and in the far-away areas, the localization is more inaccurate.
For the one-camera (NE) setup, the detection quality 
is the worst. Both FN rate and FP rate exceed $30\%$.
We find that only one camera is not able to cover
the whole roundabout.
As shown in Figure~\ref{fig:detection-distribution},
in NE only figure, the detection density at the bottom-left
area is clearly lower than other areas.
It indicates that the bottom-left area is not well-covered.
For NE+SW setup and four-camera setup, the 
coverage on the whole roundabout is good.

\begin{table}[t]
    \renewcommand{\arraystretch}{1.3}
    \setlength{\tabcolsep}{8pt}
    \centering
    \begin{tabular}{l|c|c|c}
    \shline
        \#Trip & FP rate & FDE$_{1.2s}$(Lat. error) & FDE$_{1.2s}$(Lon. error) \\
        \shline
        $1$ & $20.26$ & $0.79$ m & $1.48$ m \\
        $2$ & $2.99$ & $0.74$ m & $0.86$ m \\
        $3$ & $4.08$ & $0.62$ m & $1.73$ m \\
        $4$ & $15.79$ & $0.60$ m & $1.22$ m \\
        $5$ & $36.36$ & $0.75$ m & $1.72$ m \\
        $6$ & $19.05$ & $0.32$ m & $0.98$ m \\
        $7$ & $35.00$ & $0.84$ m & $0.75$ m \\
        $8$ & $4.55$ & $0.82$ m & $1.23$ m \\
        \hline
        Overall & $16.01$ & $0.69$ m & $1.25$ m \\
        \shline
    \end{tabular}
    \caption{Evaluations of trajectory prediction
    performance of MSight system. The False Positive rate,
    latitude error
    and longitudinal error of the predicted vehicle
    positions are reported.}
    \label{tab:prediction}
\end{table}

\subsection{Trajectory prediction results}
Table~\ref{tab:prediction} shows the
trajectory prediction results of the MSight system.
We use the four-camera setup for this experiment
to get the most accurate object detection results.
The overall False Positive rate is $16.01$, 
the FDE$_{1.2s}$(Lat. error) is $0.69$ m, and
FDE$_{1.2s}$(Lon. error) is $1.25$ m.
For the $8$ trips, the trip $\#5$ and trip $\#7$
have a much larger error than other trips.
We visualize the predicted trajectory of these two
trips and find out that, our algorithm predicts
a different path from the ground-truth path.
In trip $\#5$, the vehicle enters the roundabout
in the outer lane (not the right-turn-only lane).
Still, the vehicle performs a right turn, which
is abnormal behavior. Our prediction algorithm
predicts the vehicle to go straight instead,
which is the most common driving behavior.
For trip $\#7$, the prediction algorithm also
falsely predict the future vehicle path.

\subsection{System latency}
\begin{figure}
    \centering
    \includegraphics[width=.95\linewidth]{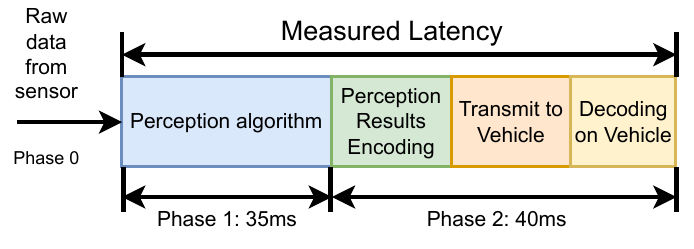}
    \caption{Latency measurement result of the system}
    \label{fig:label}
\end{figure}

Latency is a very important aspect of cooperative driving as most cooperative driving systems are delay-sensitive. To analyze the system latency clearly, we break down the cooperative perception pipeline into \textbf{three} phases:
\begin{enumerate}[start=0]
    \item \textbf{Sensor data processing}: Sensors sense and generate the data. In this phase, the sensors sense the real-world environment and provide the raw sensor data (In this case, the fisheye images), we denote this phase \textbf{phase 0}.
    \item \textbf{Perception algorithms execution}: In this phase, the perception algorithms introduced in Section \ref{section:algo} are executed, and produce final perception results. This is the core perception phase of the overall cooperative perception pipeline, we denote this phase as \textbf{phase 1}.
    \item \textbf{Perception results transmission}: In this phase, the perception results are transmitted to the vehicle from the roadside. During this phase,  the perception results are first encoded into V2X messages;
    then, V2X radio forwards the encoded messages to the vehicle; finally, on the vehicle side, the onboard unit decodes the messages. We denote this phase as \textbf{phase 2}.
\end{enumerate}
\textbf{Phase 0} is specific to sensors, which is not the focus of this paper. Consequently, in this paper, only latency in \textbf{phase 1} and \textbf{phase 2} is measured and presented. Figure \ref{fig:label} shows the breakdown of the perception pipeline and the measured latency. 

The \textbf{phase 1} latency is crucial for determining how well the core perception algorithm performs. It is measured by running the algorithm in the production environment. Phase 1's average latency is \textbf{35 ms}, which proves the effectiveness of the core algorithms. As for \textbf{phase 2}, latency is measured by timestamping each message before encoding and comparing it with the time after decoding. \textbf{Phase 2} latency is also critical as this phase is the major additional component of cooperative perception in comparison to onboard perception. The average latency of phase 2 is \textbf{40 ms}. This indicates that the roadside cooperative perception system introduced in this paper adds only \textbf{40 ms} of latency over onboard perception (if all the other components execute equally fast). This proves that the system is highly viablefor cooperative driving in terms of latency.

\section{Conclusion}
In this paper, we introduce MSight, a comprehensive roadside cooperative perception system leveraging roadside cameras, designed explicitly for applications related to autonomous driving. MSight aims to augment onboard perception systems, addressing their limitations in complex situations where their performance is often suboptimal. Subsequently, the paper outlines field tests conducted in a production environment and discusses the derived results. The findings illustrate that the system can discern vehicles with lane precision and only imposes an additional 40 ms of latency to the onboard perception pipeline. These outcomes underscore the efficacy of the roadside perception system, revealing it as a viable solution for the development of cooperative driving systems in the future.

\bibliographystyle{IEEEtran}
\bibliography{IEEEabrv,bibfile}

\section{Biography}
\begin{IEEEbiography}[{\includegraphics[width=1in,height=1.25in,clip,keepaspectratio]{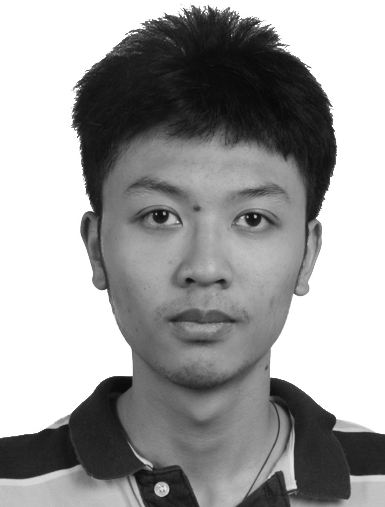}}]{Rusheng Zhang}
 received the B.E. degree in micro electrical mechanical system and second B.E. degree in Applied Mathematics from Tsinghua University, Beijing, in 2013. He received an M.S. and phD degree in electrical and computer engineering from Carnegie Mellon University, in 2015, 2019 respectively. His research areas include artificial intelligence, cooperative driving, cloud computing and vehicular networks.
\end{IEEEbiography}
\vspace{-30pt}
\begin{IEEEbiography}[{\includegraphics[width=1in,height=1.25in,clip,keepaspectratio]{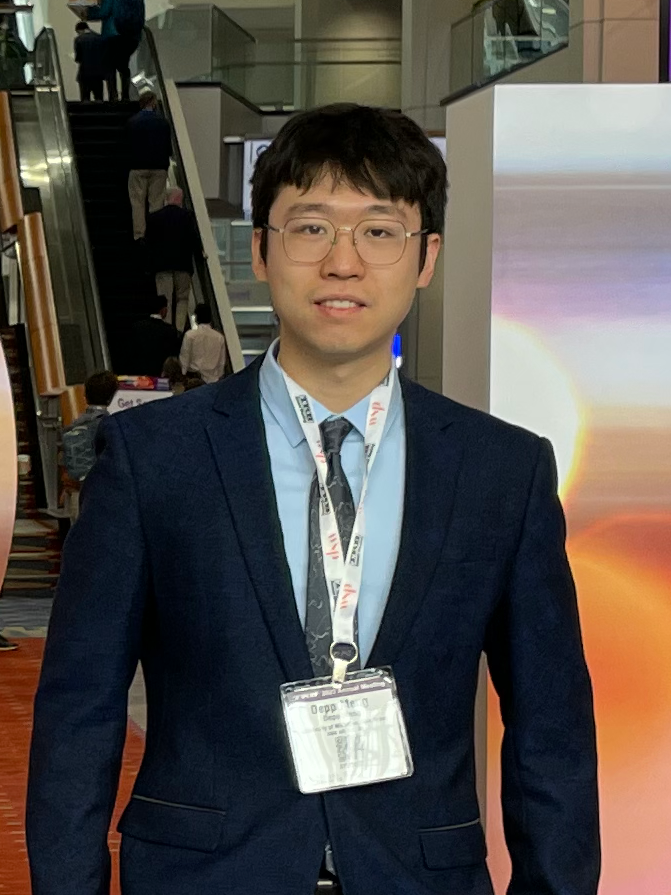}}]{Depu Meng} (Member, IEEE) is a Post Doctoral Research Fellow at the Department of Civil and Environmental Engineering, University of Michigan. He received his B. E. degree from the Department of Electrical Engineering and Information Science at the University of Science and Technology of China in 2018. He received his Ph. D. degree from the Department of Automation at the University of Science and Technology of China. His research interests include computer vision and autonomous driving systems.
\end{IEEEbiography}
\begin{IEEEbiography}[{\includegraphics[width=1in,height=1.25in,clip,keepaspectratio]{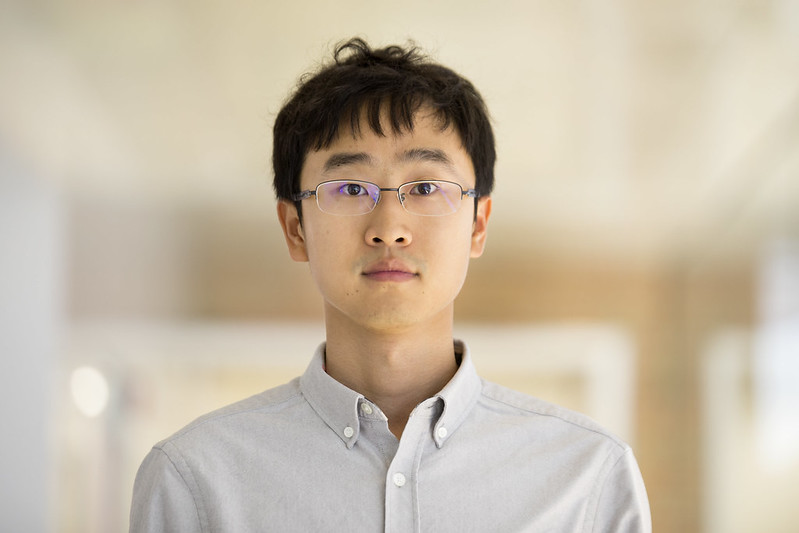}}]{Shengyin (Sean) Shen} works as a Research Engineer in the Engineering Systems Group at the University of Michigan Transportation Research Institute (UMTRI). Sean holds an MS degree in Civil and Environmental Engineering from the University of Michigan, Ann Arbor, and an MS degree in Electrical Engineering from the University of Bristol, UK. He also earned a BS degree from Beijing University of Posts and Telecommunications, China. Sean's research interests are primarily focused on cooperative driving automation and related applications that use roadside perception, edge-cloud computing, and V2X communications to accelerate the deployment of automated vehicles. He has extensive experience in implementation of large-scale deployments, such as the Safety Pilot Model Deployment (SPMD), Ann Arbor Connected Vehicle Testing Environment (AACVTE), and Smart Intersection Project. Moreover, he has been involved in many research projects funded by public agencies such as USDOT, USDOE, and companies such as Crash Avoidance Metric Partnership (CAMP), Ford Motor Company, and GM Company, among others.
\end{IEEEbiography}
\vspace{-30pt}
\begin{IEEEbiography}
[{\includegraphics[width=1in,height=1.25in,clip,keepaspectratio]{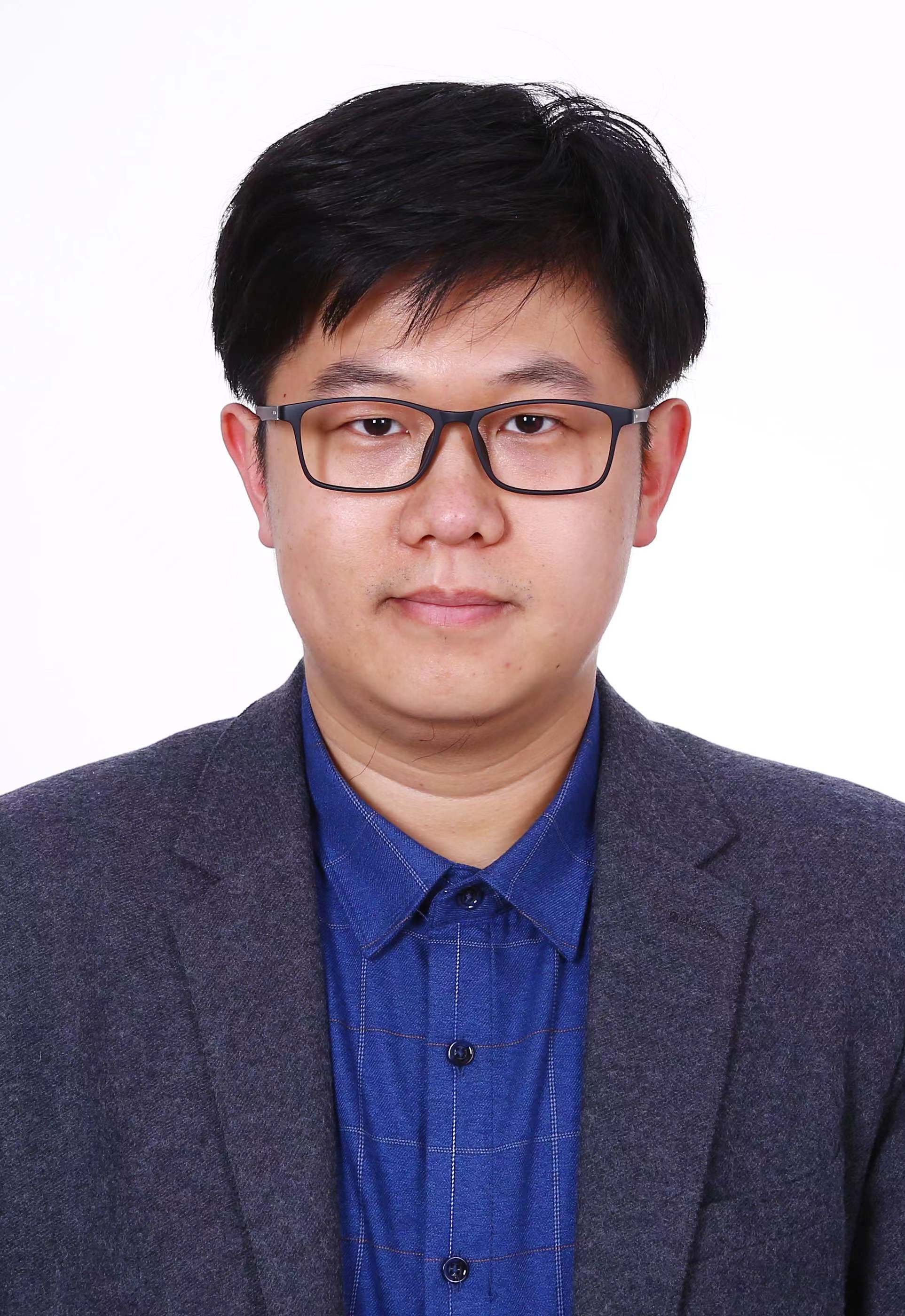}}]{Zhengxia Zou}
Zhengxia Zou received his B.S. and Ph.D. degrees from the Image Processing Center, School of Astronautics, Beihang University, Beijing, China, in 2013 and 2018, respectively. He is currently a Professor at the School of Astronautics, Beihang University. From 2018 to 2021, he worked at the University of Michigan, Ann Arbor as a Post-Doctoral Research Fellow. His research interests include computer vision and related problems in autonomous driving and remote sensing. He has published more than 40 peer-reviewed papers in top-tier journals and conferences, including Nature, Nature Communications, PROCEEDINGS OF THE IEEE, IEEE TRANSACTIONS ON IMAGE PROCESSING, IEEE TRANSACTIONS ON GEOSCIENCE AND REMOTE SENSING, and IEEE/CVF Computer Vision and Pattern Recognition. Dr. Zou was selected as “World’s Top 2\% Scientists” by Stanford University in 2022.
\end{IEEEbiography}
\vspace{-30pt}
\begin{IEEEbiography}[{\includegraphics[width=1in,height=1.25in,clip,keepaspectratio]{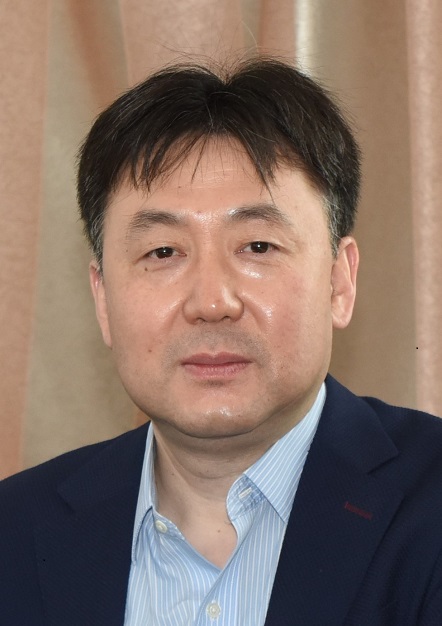}}]{Houqiang Li} (Fellow, IEEE) is a Professor with the Department of Electronic Engineering and Information Science at the University of Science and Technology of China.  His research interests include multimedia search, image/video analysis, video coding and communication. He has authored and co-authored over 200 papers in journals and conferences. He is the winner of National Science Funds (NSFC) for Distinguished Young Scientists, the Distinguished Professor of Changjiang Scholars Program of China, and the Leading Scientist of Ten Thousand Talent Program of China. He served as an Associate Editor of the IEEE Transactions on Circuits and Systems for Video Technology from 2010 to 2013. He served as the TPC Co-Chair of VCIP 2010, and he served as the General Co-Chair of ICME 2021. He is the recipient of National Technological Invention Award of China (second class) in 2019 and the recipient of National Natural Science Award of China (second class) in 2015. He was the recipient of the Best Paper Award for VCIP 2012, the recipient of the Best Paper Award for ICIMCS 2012, and the recipient of the Best Paper Award for ACM MUM in 2011. 
  
Houqiang received the B.S., M. Eng. and Ph.D degrees in electronic engineering from the University of Science and Technology of Chinae, Hefei, China in 1992, 1997 and 2000, respectively. He was elected as a Fellow of IEEE (2021).  
\end{IEEEbiography}
\vspace{-30pt}
\begin{IEEEbiography}[{\includegraphics[width=1in,height=1.25in,clip,keepaspectratio]{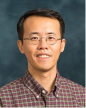}}]{Henry X. Liu}
(Member, IEEE) received the bachelor's degree in automotive engineering from Tsinghua University, China, in 1993, and the PhD. degree in civil and environment engineering from the University of Wisconsin-Madison in 2000. He is currently a professor in the Department of Civil and Environmental Engineering and the Director of Mcity at the University of Michigan, Ann Arbor. He is also a Research Professor at the University of Michigan Transportation Research Institute and the Director for the Center for Connected and Automated Transportation (USDOT Region 5 University Transportation Center). From August 2017 to August 2019, Prof. Liu served as DiDi Fellow and Chief Scientist on Smart Transportation for DiDi Global, Inc., one of the leading mobility service providers in the world. Prof. Liu conducts interdisciplinary research at the interface of transportation engineering, automotive engineering, and artificial intelligence. Specifically, his scholarly interests concern traffic flow monitoring, modeling, and control, as well as testing and evaluation of connected and automated vehicles. Prof. Liu is the managing editor of Journal of Intelligent Transportation Systems.
\end{IEEEbiography}

\vfill

\end{document}